\documentclass{article}
\usepackage{spconf}
\usepackage[colorinlistoftodos]{todonotes}
\ninept

\usepackage{amsmath,graphicx}
\usepackage[utf8]{inputenc}
\usepackage{psfrag,epsfig,graphics}
\usepackage{amsmath,amsthm,amssymb,multirow}
\usepackage{mathbbol}
\usepackage{amssymb}             

\usepackage{array}
\usepackage{mathabx} 

\DeclareSymbolFontAlphabet{\amsmathbb}{AMSb}%

\usepackage{booktabs}
\usepackage{graphicx}
\usepackage{caption}
\usepackage{subcaption}
\usepackage[position=b]{subcaption}

\usepackage[noadjust]{cite}
\usepackage{multirow}
\usepackage[lined,linesnumbered,ruled]{algorithm2e}

\usepackage{color}  

\newcommand{\cb}[1]{{\boldsymbol{#1}}}
\newcommand{\cp}[1]{\ifmmode {\mathcal{#1}}\else ${\mathcal{#1}}$\fi}

\newcommand{\bB}{\boldsymbol{B}}

\newcommand{\bI}{\boldsymbol{I}}

\newcommand{\bM}{\boldsymbol{M}}

\newcommand{\bZ}{\boldsymbol{Z}}

\newcommand{\bm}{\boldsymbol{m}}

\newcommand{\be}{\boldsymbol{e}}

\newcommand{\br}{\boldsymbol{r}}

\newcommand{\bz}{\boldsymbol{z}}

\newcommand{\tensor}[1]{\cb{\mathcal{#1}}}

\newcommand{\balpha}{\boldsymbol{\alpha}}

\title{A Low-rank Tensor Regularization Strategy\\ for Hyperspectral Unmixing }

\name{Tales Imbiriba, Ricardo~Augusto~Borsoi, Jos\'e~Carlos~Moreira~Bermudez
\thanks{This work has been supported by the National Council for Scientific and Technological Development (CNPq).}
\thanks{T. Imbiriba, R. A. Borsoi and J.C.M. Bermudez are with the Department of Electrical Engineering, Federal University of Santa Catarina, Florian\'opolis, SC, Brazil. e-mail: talesim@gmail.com; raborsoi@ucs.br; j.bermudez@ieee.org.}%
}
\address{Federal University of Santa Catarina, Florianópolis, SC, Brazil}

\begin{document}
%
\maketitle
\begin{abstract}
Tensor-based methods have recently emerged as a more natural and effective formulation to address many problems in hyperspectral imaging. In hyperspectral unmixing (HU), low-rank constraints on the abundance maps have been shown to act as a regularization which adequately accounts for the multidimensional structure of the underlying signal. However, imposing a strict low-rank constraint for the abundance maps does not seem to be adequate, as important information that may be required to represent fine scale abundance behavior may be discarded. This paper introduces a new low-rank tensor regularization that adequately captures the low-rank structure underlying the abundance maps without hindering the flexibility of the solution. Simulation results with synthetic and real data show that the the extra flexibility introduced by the proposed regularization significantly improves the unmixing results.
\end{abstract}
\section{Introduction}
\label{sec:intro}

Hyperspectral imaging has attracted formidable interest of the scientific community, and has been applied to an increasing number of applications in different fields~\cite{Bioucas-Dias-2013-ID307}. The limited spatial resolution of hyperspectral devices often mixes the spectral contribution of different pure materials, termed \emph{endmembers}, in the scene~\cite{Keshava:2002p5667}. This phenomenon is especially relevant in remote sensing applications due to the distance between airborne or spaceborne sensors and the target scene. Such mixing process must be well understood to accurately unveil vital information relating the pure materials and their distribution in the scene. Hyperspectral unmixing (HU) aims to solve this problem by factorizing the hyperspectral image (HI) into a collection of endmembers and their fractional \emph{abundances}~\cite{Bioucas2012}. 

Different mixing models have been used to explain the interaction between light and the endmembers~\cite{Dobigeon-2014-ID322,Imbiriba2016_tip, Imbiriba2017_bs_tip} or spectral variability along the image~\cite{Zare-2014-ID324,imbiriba2017GLMM}. 
%
The simplest and most widely used model is the Linear Mixing Model (LMM)~\cite{Keshava:2002p5667}, which assumes that the observed reflectance vector (\emph{i.e.} a pixel) can be modeled as a convex combination of the spectral signatures of the endmembers present in the scene. The LMM imposes positivity and sum-to-one constraints on the linear combination coefficients, which are then interpreted as the proportional contribution of each endmember to the scene (fractional endmember abundance). The simplicity of the LMM and the convexity constraints over the fractional abundances naturally lead to fast and reliable unmixing strategies. 
Nevertheless, endmember spectral signatures are frequently highly correlated. This makes the HU problem ill-posed and sensitive to the influence of the observation noise, what results in large reconstruction errors for the abundance estimates~\cite{meer2012collinearityEndmembersUnmixing}.

Significant efforts have been devoted to overcome this problem by introducing appropriate \textit{a priori} information into the HU problem. This is usually done by means of a spatial regularization, which introduces an additional constraint forcing neighboring pixels to have similar abundances~\cite{meer2012collinearityEndmembersUnmixing,shi2014surveySpatialRegUnmixing,borsoi2017tech_arxiv}. An important interpretation of spatial regularization is that it enforces the preservation of lower dimensional structures existing in the HI despite the presence of noise and other non-modeled phenomena. This rationale has led to the proposition of new efficient HI unmixing methods. For instance, in~\cite{borsoi2017tech_arxiv}, a multi-scale approach was considered to impose lower-dimensional structures in the abundance estimation. 

Possible ways to recover lower-dimensional structures from noisy and corrupted data include the imposition of low-rank matrix constraints on the estimation process~\cite{tao2011recovering}, or the low-rank decomposition of the observed data~\cite{bousse2017tensor,sidiropoulos2017tensor}. 
The facts that HIs are naturally represented and treated as tensors, and that low-rank decompositions of higher-order ($>$2) tensors tend to capture homogeneities within the tensor structure make the latter strategies even more attractive for HU.
Low-rank tensor models have been successfully employed in various tasks involving HIs, such as recovery of missing pixels~\cite{ng2017tensorHSIrecoveryMissing}, anomaly detection~\cite{zhang2016tensorAnomalyDetectionHSI}, classification~\cite{guo2016supportTensorMachines}, compression~\cite{yang2015tensorHSIcompression}, dimensionality reduction~\cite{zhang2013tensorDimensionalityReductionHSI} and analysis of multi-angle images~\cite{veganzones2016tensorCPdecHSImultiangle}.
More recently, Qian \emph{et al}~\cite{qian2017tensorNMFunmixing} considered a low-rank decomposition of HIs for solving the HU problem using a nonnegative tensor factorization (NTF) strategy where the spatial regularity is enforced through the imposition of a low-rank tensor structure. Though a low-rank tensor representation may naturally describe the regularity of HIs and abundance maps, the forceful introduction of stringent rank constraints may prevent an adequate representation of important fast varying structures. 
%
Another limitation of the NTF approach in~\cite{qian2017tensorNMFunmixing} is the lack of guarantee that endmembers and abundances will be correctly factorized in their respective tensors.

In this work we propose a new low-rank method for HU that accounts for highly correlated endmembers, called \textit{Unmixing with Low-rank Tensor Regularization Algorithm} (ULTRA).  We formulate the HU problem using tensors and introduce a low-rank abundance tensor regularization term.  This strategy allows important flexibility to the rank of the estimated abundance tensor to adequately represent fine scale structure and details which lie beyond a low-rank structure, but without compromising the smoothness of the solution. The proposed HU method is supervised, in that the endmember matrix is assumed to be known \emph{a priori} or estimated using a endmember extraction algorithm such as the VCA~\cite{Nascimento2005}.   
%
%
Our experiments indicate that the proposed approach represents a promising alternative for imposing spatial regularity on the unmixing problem.   

This paper is organized as follows. In Section~\ref{sec:back_notation} we introduce the notation and the relevant tensor decomposition. In Section~\ref{sec:prop_tensor_HU} we present the tensor formulation for HU, and the proposed regularized unmixing problem. We then compare the performance of the proposed method with those of competing algorithms in Section~\ref{sec:Simulations}. Finally, we present conclusions in Section~\ref{sec:conclusions}.


\subsection{The Linear Mixing Model}\label{sec:LMMs}

The Linear Mixing Model (LMM)~\cite{Keshava:2002p5667} assumes that a given pixel $\br_{n_1,n_2} = [r_{n_1,n_2,1},\,\ldots, \,r_{n_1,n_2,L} ]^\top$ with $L$ bands is represented as
\begin{equation}
\begin{split}
 &\br_{n_1,n_2} = \bM \balpha_{n_1,n_2} + \be_{n_1,n_2}\\
 &\text{subject to }\,\cb{1}^\top\balpha_{n_1,n_2} = 1 \text{ and } \balpha_{n_1,n_2} \succeq \cb{0}
\end{split}
\label{eq:LMM}
\end{equation}
where $\bM = [\bm_1,\,\ldots, \,\bm_R]$ is an $L\times R$ matrix composed of the $R$ endmember spectral signatures $\bm_i = [m_{i,1},\,\ldots,\,m_{i,L}]^\top$, $\balpha_{n_1,n_2} = [\alpha_{n_1,n_2,1},\,\ldots,\,\alpha_{n_1,n_2,R}]^\top$ is the abundance vector, $\be_{n_1,n_2}\sim\mathcal{N}(0, \sigma_{n_1,n_2}^2\bI_L)$ is an additive white Gaussian noise (WGN), $\bI_L$ is the $L\times L$ identity matrix, and $\succeq$ is the entry-wise $\geq$ operator.

\begin{figure}
\centering
\includegraphics[width=0.47\textwidth]{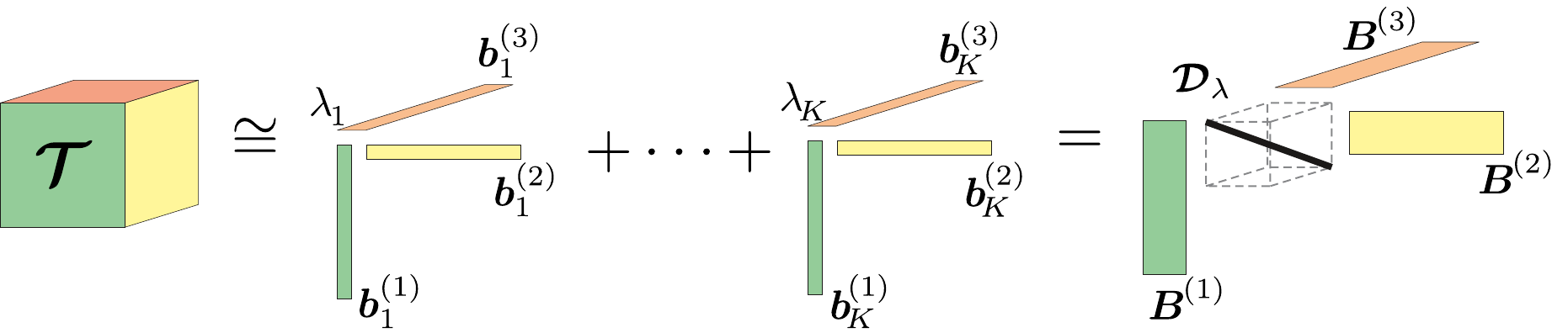}
\caption{Polyadic decomposition of a three-dimensional tensor, written as both outer products and mode$-n$ products.}
\label{fig:tensor_illustr}
\end{figure}

\section{Background and notation}
\label{sec:back_notation}

\subsection{Notation}
An order-$P$ tensor $\tensor{T} \in \amsmathbb{R}^{N_1\times \dots \times N_P}$ ($P>2$) is an $N_1\times \dots \times N_P$ array  with elements indexed by~$\tensor{T}_{n_1,n_2,\ldots,n_P}$. The $P$ dimensions of a tensor are called \textit{modes}. A mode-$\ell$ fiber of tensor $\tensor{T}$ is the one-dimensional subset of  $\tensor{T}$ obtained by fixing all but the $\ell$-th dimension, and is indexed by $\tensor{T}_{n_1,\ldots,n_{\ell-1},:,n_{\ell+1},\ldots,n_P}$. A slab or slice of tensor $\tensor{T}$ is a two-dimensional subset of $\tensor{T}$ obtained by fixing all but two of its modes.
An HI is often conceived as a three dimensional data cube, and can be naturally represented by an order-3 tensor $\tensor{R}\in\amsmathbb{R}^{N_1\times N_2\times L}$, containing $N_1\times N_2$ pixels represented by the tensor fibers $\tensor{R}_{n_1,n_2,:}\in\amsmathbb{R}^{L}$. 
Analogously, the abundances can also be collected in an order-3 tensor $\tensor{A}\in\amsmathbb{R}^{N_1\times N_2\times R}$. Thus, given a pixel $\tensor{R}_{n_1,n_2,:}$, the respective abundance vector $\balpha_{n_1,n_2}$ is represented by the mode-3 fiber $\tensor{A}_{n_1,n_2,:}$.
We now review some operations of multilinear algebra (the algebra of tensors) that will be used in the following sections (more details can be found in~\cite{cichocki2015tensor}).


\subsection{Tensor product definitions}
\textbf{Outer product:}
The outer product between vectors $\cb{b}^{(1)}\in\amsmathbb{R}^{N_1},\cb{b}^{(2)}\in\amsmathbb{R}^{N_2},\ldots,\cb{b}^{(P)}\in\amsmathbb{R}^{N_P}$ is defined as the order-$P$ tensor $\tensor{T}=\cb{b}^{(1)}\circ\cb{b}^{(2)}\circ\cdots\circ\cb{b}^{(P)}\in\amsmathbb{R}^{N_1\times N_2\times\cdots\times N_P}$, where $\tensor{T}_{n_1,n_2,\ldots,n_P}=b^{(1)}_{n_1}b^{(2)}_{n_2}\cdots b^{(P)}_{n_P}$ and $b^{(i)}_{n_i}$ is the $n_i$-th position of $\cb{b}^{(i)}$. It generalizes the outer product between two vectors.


\noindent\textbf{Mode-$k$ product:}
The mode-$k$ product, denoted $\tensor{U}=\tensor{T}\times_k\cb{B}$, of a tensor $\tensor{T} \in \amsmathbb{R}^{N_1\times \dots\times N_k\times \dots \times N_P}$ and a matrix $\cb{B} \in \amsmathbb{R}^{M_k\times N_k}$ is evaluated such that each mode-$k$ fiber of $\tensor{T}$ is multiplied by matrix $\cb{B}$, yielding $\tensor{U}_{n_1,n_2,\ldots,m_k,\ldots,n_P}=\sum_{i=1}^{N_k}\tensor{T}_{\ldots,n_{n-1},i,n_{n+1},\ldots} \cb{B}_{m_k,i}$.


\noindent\textbf{Multilinear product:}
The full multilinear product, denoted by $\big\ldbrack\tensor{T};\bB^{(1)},\bB^{(2)},\ldots,\bB^{(P)}\big\rdbrack$, consists of the successive application of mode-$k$ products between $\tensor{T}$ and matrices $\bB^{(i)}$, represented as $\tensor{T}\times_1\bB^{(1)}\times_2\bB^{(2)}\times_3\ldots\times_P\bB^{(P)}$.

\noindent\textbf{Mode-(M,1) contracted product:}
The contracted mode-M product, denoted by $\tensor{U}=\tensor{T}\times^M\cb{b}$, is a product between a tensor $\tensor{T}$ and a vector $\cb{b}$ in mode-M, where the resulting singleton dimension is removed, given by $\tensor{U}_{\ldots,n_{n-1},n_{n+1},\ldots}=\sum_{i=1}^{N_n}\tensor{T}_{\ldots,n_{n-1},i,n_{n+1},\ldots}\cb{b}_i$.

\subsection{The Canonical Polyadic Decomposition}
An order-$P$ rank-1 tensor is obtained as the outer product of $P$ vectors. The rank of an order-$P$ tensor $\tensor{T}$ is defined as the minimum number of order-$P$ rank-1 tensors that must be added to obtain $\tensor{T}$~\cite{sidiropoulos2017tensor}. Thus, any tensor $\tensor{T}\in\amsmathbb{R}^{N_1\times N_2\times\cdots\times N_P}$ with $\text{rank}(\tensor{T})=K$ can be decomposed as a linear combination of at least $K$ outer products of $P$ rank-1 tensors. This so-called \textit{polyadic decomposition} is illustrated in Figure~\ref{fig:tensor_illustr}. When this decomposition involves exactly $K$ terms, it is called the canonical polyadic decomposition  (CPD)~\cite{cichocki2015tensor} of a rank-$K$ tensor $\tensor{T}$, and is given by

%
%
\begin{equation}
\tensor{T} = \sum_{i=1}^{K} \lambda_i\cb{b}^{(1)}_i\circ\cb{b}^{(2)}_i\circ\cdots\circ \cb{b}^{(P)}_i.
\end{equation}
It has been shown that this decomposition is essentially unique under mild conditions~\cite{sidiropoulos2017tensor}. 
The CPD can be written alternatively using mode-$k$ products as
\begin{equation}
\tensor{T} = \tensor{D}_\lambda\times_1 \bB^{(1)}\times_2 \bB^{(2)}\cdots \times_{\!P} \bB^{(P)}
\end{equation}
or using the  full multilinear product as
\begin{equation}
\tensor{T} = \big\ldbrack \tensor{D}_\lambda;\bB^{(1)},\bB^{(2)},\ldots,\bB^{(P)} \big\rdbrack
\end{equation}
where $\tensor{D}_\lambda = \text{Diag}_P\big(\lambda_{1},\ldots,\lambda_{K}\big)$ is the $P$-dimensional diagonal tensor. 
Given a tensor $\tensor{T}\in\amsmathbb{R}^{N_1\times N_2\times\cdots\times N_P}$, the CPD can be obtained by solving the following optimization problem~\cite{sidiropoulos2017tensor}
\begin{equation}
\begin{split}
 \Big( \widehat{\tensor{D}}_\lambda,&\widehat{\bB}^{(1)},\ldots,\widehat{\bB}^{(K)}\Big) =
 \\&
 \mathop{\arg\min}_{\tensor{D}_\lambda,\bB^{(1)},\ldots,\bB^{(K)}}\frac{1}{2}\Big\|\tensor{T} - \sum_{i=1}^{K} \lambda_{i}\cb{b}_i^{(1)}\circ\cdots\circ\cb{b}_i^{(P)}\Big\|^2_F 
 .
\end{split}
\label{eq:cpd_opt_i}
\end{equation}


A widely used strategy to compute an approximate solution to~\eqref{eq:cpd_opt_i} is to use an alternating least-squares technique~\cite{sidiropoulos2017tensor}, which optimizes the cost function with respect to one term at a time, while keeping the others fixed, until convergence. Although optimization problem~\eqref{eq:cpd_opt_i} is generally non-convex, its solution is unique under relatively mild conditions, which is an important advantage of tensor-based methods~\cite{sidiropoulos2017tensor}.


\section{The low-rank unmixing problem}
\label{sec:prop_tensor_HU}

%
A strategy to solve the HU problem that can effectively capture the low-dimensional structures is to impose a low-rank to the abundance tensor~\cite{qian2017tensorNMFunmixing}. 
%
Thus, assuming that $\tensor{A}$ has a low-rank $K_\tensor{A}$, the global cost functional for the unmixing problem can be written as
\begin{equation}
\begin{split}
 J(\tensor{A}&) = \frac{1}{2}\sum_{n_1=1}^{N_1}\sum_{n_2=1}^{N_2}\|\tensor{R}_{n_1,n_2,:} -\bM\tensor{A}_{n_1,n_2,:} \|_F^2 \\
  \text{s. t.}\;&\; \text{rank}(\tensor{A})= K_\tensor{A},\,\tensor{A}\succeq \cb{0},\, \tensor{A}\times^1 \cb{1}_{R} = \cb{1}_{N_1\times N_2}.  
\end{split}\label{eq:unmixing_cost_func}
\end{equation}
This cost function, however, imposes a very strict and data independent condition on the rank of $\tensor{A}$, limiting its flexibility to adequately represent the desired abundance maps.
Although fixing a low-rank for $\tensor{A}$ tends to capture the most significant part of the abundance matrix energy~\cite{Mei_lowrank_2018}, one may incur in a loss of fine and small scale details that may be important for specific data.
On the other hand, using a large value for $K_{\tensor{A}}$ makes the solution sensitive to noise, undermining the purpose of a regularization.
Hence, an important issue is how to effectively impose the low-rank constraint to achieve regularity in the solution without undermining its flexibility to adequately model small variations and details.

To deal with this issue we propose to introduce a new regularization term controlled by a low-rank tensor $\tensor{Q}\in\amsmathbb{R}^{N_1\times N_2\times R}$ with the purpose of providing a non-strict constraint on $K_{\tensor{A}}$. Doing that, tensor $\tensor{Q}$ works as an  \emph{a prior} information, and the strictness of the low-rank constraint is controlled by an additional parameter $\lambda_\tensor{A}\in \amsmathbb{R}^+$. Thus, we write the proposed alternative optimization problem as
\begin{equation}
\begin{split}
 J(\tensor{A},\tensor{Q}) {}={}& \frac{1}{2}\sum_{n_1=1}^{N_1}\sum_{n_2=1}^{N_2}\|\tensor{R}_{n_1,n_2,:} -\bM\tensor{A}_{n_1,n_2,:} \|_F^2 \\
  & + \frac{\lambda_\tensor{A}}{2}\|\tensor{A}-\tensor{Q}\|_F^2\\
  \text{s. t.}\;&\; \tensor{A}\succeq \cb{0},\, \tensor{A}\times^{\!1} \cb{1}_{R} = \cb{1}_{N_1\times N_2}
\end{split}\label{eq:unmixing_cost_func2}
\end{equation}
where $\tensor{Q}$ is a tensor with $\text{rank}(\tensor{Q}) = K_\tensor{Q}$.

Thus, the optimization problem becomes
%
\begin{equation}
(\,\widehat{\!\!\tensor{A}},\widehat{\tensor{Q}}) = \mathop{\arg\min}_{\tensor{A},\tensor{Q}} J(\tensor{A},\tensor{Q}).
\label{eq:opt_tensor}
\end{equation}
%
We propose to find a local stationary point minimizing~\eqref{eq:opt_tensor} iteratively with respect to each variable, leading to the \textit{Unmixing with Low-rank Tensor Regularization Algorithm} (ULTRA) presented in $\text{Algorithm}$~1. The intermediate steps are detailed in the following.

\begin{algorithm} [bth]
\footnotesize
\SetKwInOut{Input}{Input}
\SetKwInOut{Output}{Output}
\caption{Unmixing with Low-rank Tensor Regularization Algorithm (ULTRA)~\label{alg:global_opt2}}
\Input{$\tensor{R}$, $\lambda_\tensor{A}$, $K_\tensor{Q}$, $\tensor{A}^{(0)}$,$\bM$, and $\tensor{Q}^{(0)}$.}
\Output{$\,\,\widehat{\!\!\tensor{A}}$ and $\,\widehat{\!\tensor{Q}\!}$.}
Set $i=0$ \;
\While{stopping criterion is not satisfied}{
$i=i+1$ \;
$\tensor{A}^{(i)} = \underset{\tensor{A}}{\arg\min} \,\,\,\,  {J}(\tensor{A},\tensor{Q}^{(i-1)})$ \;
$\tensor{Q}^{(i)} = \underset{\tensor{P}}{\arg\min} \,\,\,\,  {J}(\tensor{A}^{(i)},\tensor{Q})$ \;
}
\KwRet $\,\,\widehat{\!\!\tensor{A}}=\tensor{A}^{(i)}$,~ $\,\widehat{\!\tensor{Q}\!}=\tensor{Q}^{(i)}$  \;
\end{algorithm}

\subsection{Solving with respect to $\tensor{A}$}
To solve problem~\eqref{eq:opt_tensor} with respect to the abundance tensor $\tensor{A}$ we can rewrite the terms in the cost functional~\eqref{eq:unmixing_cost_func2} that depend on $\tensor{A}$ independently for each pixel, leading to
\begin{equation}
\begin{split}
 J(\tensor{A}) {}={} & \frac{1}{2}\sum_{n_1=1}^{N_1}\sum_{n_2=1}^{N_2}\|\tensor{R}_{n_1,n_2,:} -\bM\tensor{A}_{n_1,n_2,:} \|_F^2 \\
  & + \frac{\lambda_\tensor{A}}{2}
  \sum_{n_1=1}^{N_1}\sum_{n_2=1}^{N_2}
  \|\tensor{A}_{n_1,n_1,:} - \tensor{Q}_{n_1,n_2,:}\|^2_F \\
  \text{s. t.}\;&\; \tensor{A}\succeq \cb{0},\, \tensor{A}\times^1 \cb{1}_{R} = \cb{1}_{N_1\times N_2}.
\end{split}
\end{equation}
which is a standard regularized fully constrained least-squares problem and can be solved efficiently.

\subsection{Solving with respect to $\tensor{Q}$}
Analogous to the previous section, the cost function to be optimized for $\tensor{Q}$ can be written as
\begin{equation}
 J(\tensor{Q}) = \frac{\lambda_\tensor{A}}{2}\|\tensor{A} - \tensor{Q}\|^2_F 
 \label{eq:Q_problem}
\end{equation}

Assuming that most of the energy of $\tensor{A}$ lies in a low-rank structure, we write tensor $\tensor{Q}$ as a sum of a small number $K_\tensor{Q}$ of rank-1 components, such that
\begin{equation}
 	\tensor{Q} = \sum_{i=1}^{K_\tensor{Q}} 		
    \xi_{i}\bz^{(1)}_i\circ\bz^{(2)}_{i}\circ\bz^{(3)}_{i}.
 \label{eq:Q_cpd}
\end{equation}
This introduces a low-rank \emph{a priori} condition for $\tensor{A}$, which will be more or less enforced depending on the regularization constant $\lambda_\tensor{A}$. 
Replacing~\eqref{eq:Q_cpd} in~\eqref{eq:Q_problem} leads to the following optimization problem
\begin{equation}
\begin{split}\label{eq:Q_als}
 \Big(\widehat{\cb{\Xi}}\,\,,&\,\,\widehat{\bZ}^{(1)}\!,\widehat{\bZ}^{(2)}\!,\widehat{\bZ}^{(3)}\Big) = \\ 
 & \mathop{\arg\min}_{\cb{\Xi},\bZ^{(1)},\bZ^{(2)},\bZ^{(3)}}\frac{\lambda_\tensor{A}}{2}\Big\|\tensor{A} - \sum_{i=1}^{K_\tensor{Q}} \xi_{i}\bz_i^{(1)}\circ\bz_i^{(2)}\circ\bz_i^{(3)}\Big\|^2_F
\end{split}
\end{equation} 
where $\cb{\Xi} = \text{Diag}_3\big(\xi_{1}, \ldots, \xi_{K_{\!\tensor{Q}}}\big)$ is an order-3 diagonal tensor with $\cb{\Xi}_{i,i,i}=\xi_i$.
Problem~\eqref{eq:Q_als} can be solved using an alternating least-squares strategy~\cite{sidiropoulos2017tensor}.
Finally, the solution $\widehat{\tensor{Q}}$ is obtained from $\widehat{\cb{\Xi}},\widehat{\bZ}^{(1)},\widehat{\bZ}^{(2)}$ and $\widehat{\bZ}^{(3)}$ by using the full multilinear product as
\begin{equation}
 \widehat{\tensor{Q}} = \big\ldbrack \widehat{\cb{\Xi}} \,\,;\, \widehat{\bZ}^{(1)}; \widehat{\bZ}^{(2)};\widehat{\bZ}^{(3)} \big\rdbrack.
\end{equation}

\begin{table}[htb!]
\caption{Results for simulations with synthetic data.}
\begin{center}
\renewcommand{\arraystretch}{1}
\vspace{-0.5cm}
\begin{tabular}{m{1.1cm}| c m{0.75cm}| c m{0.75cm}}
\toprule
\multicolumn{5}{c}{Data Cube 0 -- DC0}\\
\bottomrule\toprule
& \multicolumn{2}{c|}{25 dB} & \multicolumn{2}{c}{15 dB}\\
\midrule
& $\text{SRE}_{\tensor{A}}$ & Time & $\text{SRE}_{\tensor{A}}$ & Time \\
\midrule
FCLS	&	20.17	$\pm$	1.25	&	0.4	&	10.86	$\pm$	1.14	&	0.4	\\
NTF	&	3.29	$\pm$	0.47	&	59.5	&	3.35	$\pm$	0.55	&	41.3	\\
ULTRA	&	\textbf{21.09	$\pm$	1.51}	&	3.0	&	\textbf{11.78	$\pm$	1.35}	&	1.5	\\
\bottomrule\toprule
\multicolumn{5}{c}{Data Cube 1 -- DC1}\\
\bottomrule\toprule
FCLS	&	10.67	$\pm$	1.52	&	1.1	&	3.99	$\pm$	1.17	&	1.0	\\
NTF	&	5.60	$\pm$	0.13	&	403.4	&	5.68	$\pm$	0.21	&	204.0	\\
ULTRA	&	\textbf{13.70	$\pm$	2.44}	&	1.4	&	\textbf{7.63	$\pm$	2.68}	&	1.0	\\
\bottomrule\toprule
\multicolumn{5}{c}{Data Cube 2 -- DC2}\\
\bottomrule\toprule
FCLS	&	16.06	$\pm$	0.92	&	0.4	&	11.17	$\pm$	0.59	&	0.4	\\
NTF	&	4.25	$\pm$	0.26	&	93.1	&	6.49	$\pm$	0.09	&	170.3	\\
ULTRA	&	\textbf{17.54	$\pm$	1.27}	&	3.7	&	\textbf{12.23	$\pm$	0.76}	&	1.5	\\
\bottomrule\toprule
\end{tabular}
\end{center}
\vspace{-0.5cm}
\label{tab:results_synthData}
\end{table}

\begin{figure}[htb]
\centering
\includegraphics[width=0.12\textwidth]{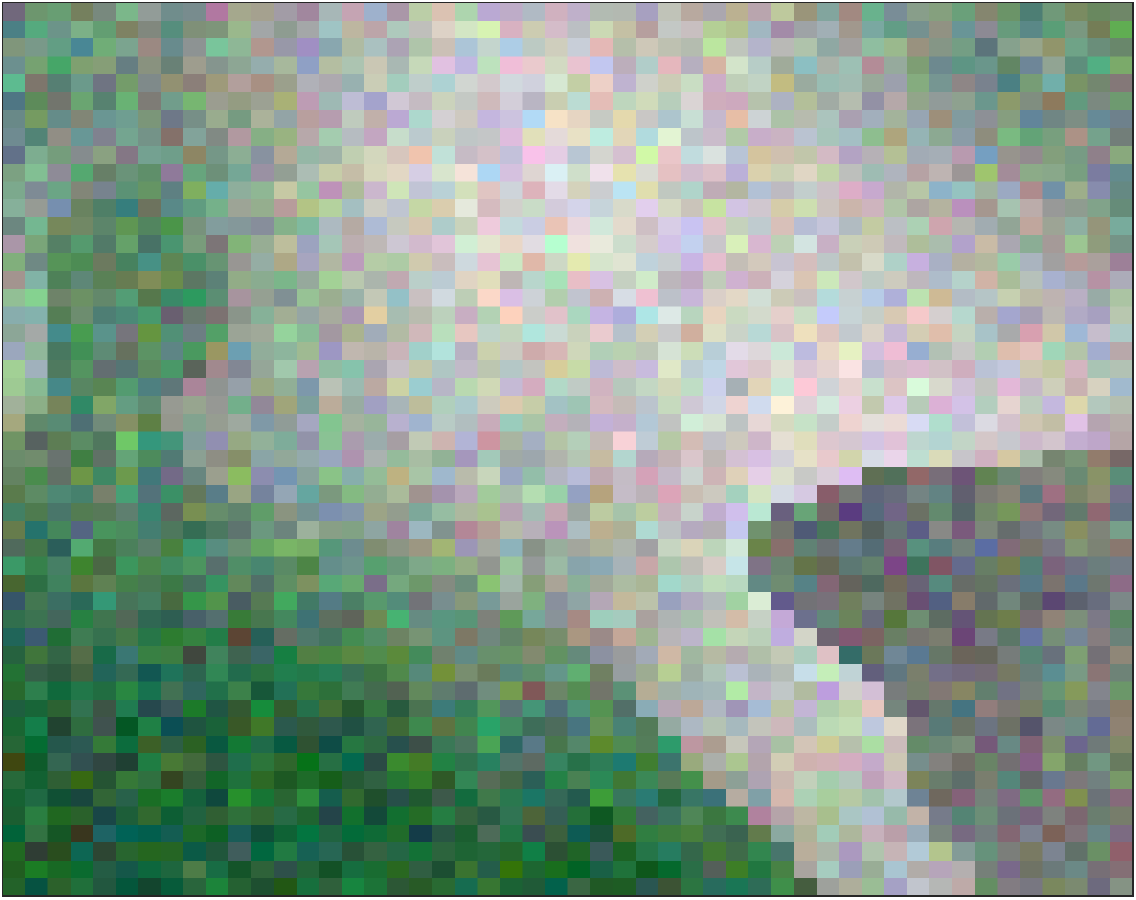}~~
\includegraphics[width=0.12\textwidth]{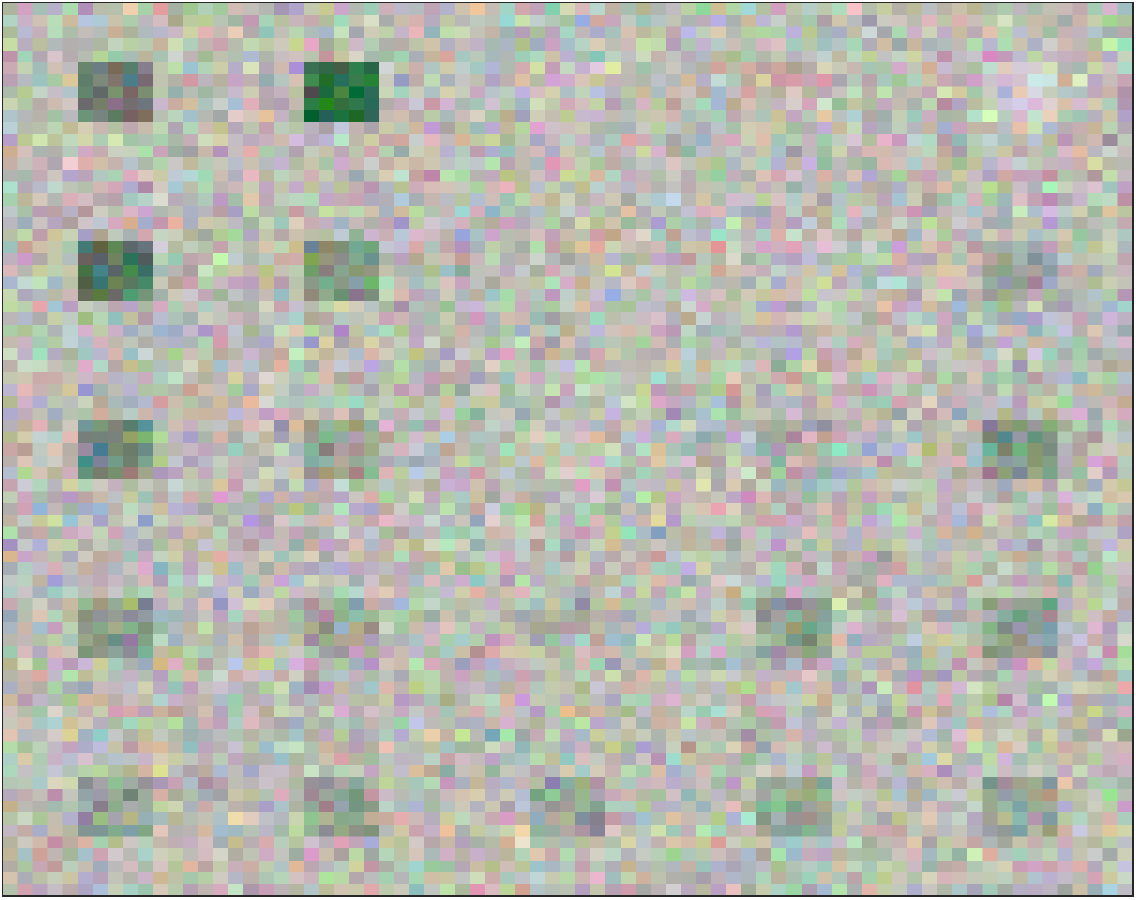}~~
\includegraphics[width=0.12\textwidth]{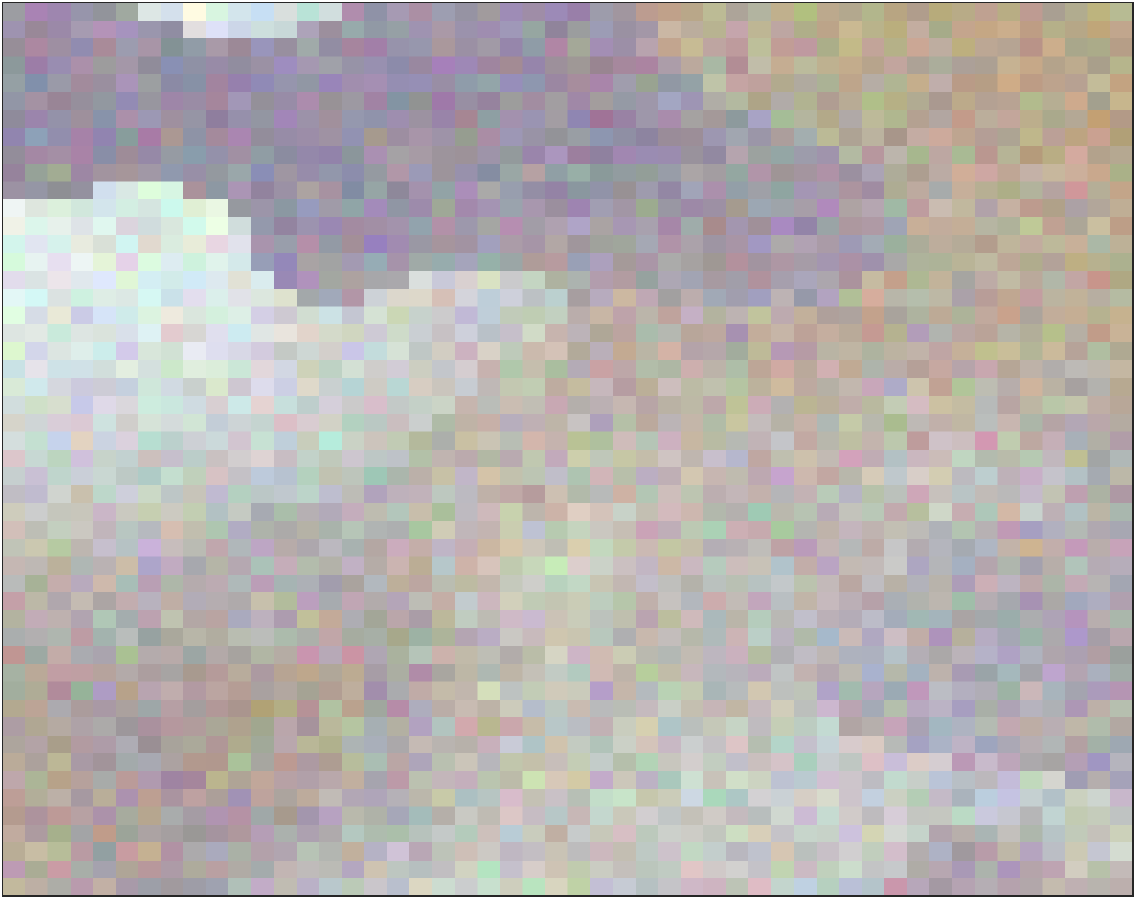}
\caption{Synthetic data cubes DC0, left, DC1, middle, and DC2, right. }\label{fig:synthDatas}
\end{figure}

\begin{figure}[htb]
\centering
\includegraphics[width=0.4\textwidth]{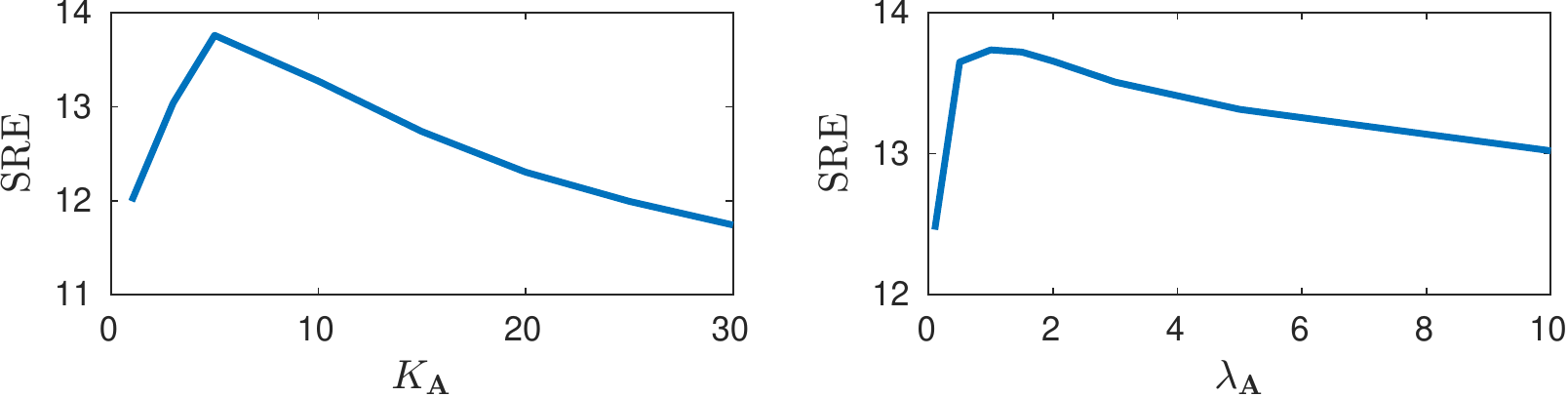}
\caption{SRE for DC1 (25dB) as a function of $K_{\tensor{A}}$ and $\lambda_{\tensor{A}}$.}
\label{fig:parSensibility}
\end{figure}

\section{Simulations}\label{sec:Simulations}
This section illustrates the performance of the proposed ULTRA through simulations with both synthetic and real data. We compare ULTRA with the the fully constrained least squares (FCLS) and with the NTF method~\cite{qian2017tensorNMFunmixing}, which is based on the problem~\eqref{eq:unmixing_cost_func}.




We measure the accuracy of the unmixing methods using the abundance signal reconstruction error (SRE) defined as
\begin{equation} \label{eq:sre_metric}
	\text{SRE}_{\tensor{A}} = 10 \log_{10}
    \bigg(\frac{\|\tensor{A}\|_F^2}
    {\|\tensor{A}-\widehat{\tensor{A}}\|_F^2} \bigg)
    \,\text{.}
\end{equation}

\subsection{Synthetic data}
For a comprehensive comparison among the different methods we created three synthetic datasets, namely, Data Cube 0 (DC0), Data Cube 1 (DC1) and Data Cube 2 (DC2), represented in Fig.~\ref{fig:synthDatas}. These datasets were built using correlated endmembers extracted from the USGS Spectral Library~\cite{clark2003imaging}, and different strategies were used to generate the abundance maps exhibiting spatial correlation between neighboring pixels.
White Gaussian noise with SNRs of 15dB or 25dB was later added to each dataset, resulting in two instances of each data cube.

To find the optimal parameters for the selected algorithms we performed a grid search for each dataset. The parameter range for the NTF method was selected as proposed by the authors in~\cite{qian2017tensorNMFunmixing}. Specifically, we fixed $\delta=0.4$ and varied the rank of the abundance matrix in the range $[5,\, 60]$. For the ULTRA method, the parameter search occurred in the intervals $[0.1,\, 10]$ for $\lambda_{\tensor{A}}$ and $[5,\, 30]$ for $K_{\tensor{Q}}$.
For instance, the optimal values for DC1 with 25dB SNR were found to be $K_\tensor{Q} = 5$ and $\lambda_\tensor{A} = 1$.
Table~\ref{tab:results_synthData} shows the results obtained using the three methods averaged over 30 realizations.
It can be verified that the ULTRA clearly outperformed the competing algorithms for all datasets and SNRs, performing significantly better than the NTF algorithm and better than FCLS. These initial results indicate that the extra flexibility provided by the proposed regularization is beneficial for tensor formulations of the HU problem.
To verify the statistical significance of the results shown in Table~\ref{tab:results_synthData}, we performed the one-tailed left nonparametric Wilcoxon signed rank test~\cite{marques2003applied} between the SREs obtained by the  ULTRA and the FCLS algorithm. In all cases the null hypothesis, $\text{median}(\text{SRE}_{\text{FCLS}}) - \text{median}(\text{SRE}_{\text{ULTRA}}) = 0$, was rejected, i.e., there was enough evidence that $\text{median}(\text{SRE}_{\text{ULTRA}}) > \text{median}(\text{SRE}_{\text{FCLS}})$, at the 0.05 significance level.

To test the sensitivity of the ULTRA performance with respect to choice of the parameters $\lambda_\tensor{A}$ and $K_\tensor{Q}$, we performed a simulation fixing one of the parameters at a time at its optimal value and varying the other. The obtained values of $\text{SRE}_{\tensor{A}}$ for DC1 are shown in Fig.~\ref{fig:parSensibility}. Although it is clear that the results degraded as the parameter moved away from their optimal values, the SRE obtained was still considerably higher than the value of 10.64 obtained by the FCLS for this specific execution or the 10.67 average behavior presented in Table~\ref{tab:results_synthData}.

The average execution time for the ULTRA method was 3 times larger than for the FCLS, but 80 times smaller than for the NTF algorithm. The significant difference between the ULTRA method and the NTF execution times is because each subproblem of the proposed optimization problem is amenable to efficient solutions.

\begin{figure}[htb]
\centering
\includegraphics[height=0.45\textwidth, angle =-90]{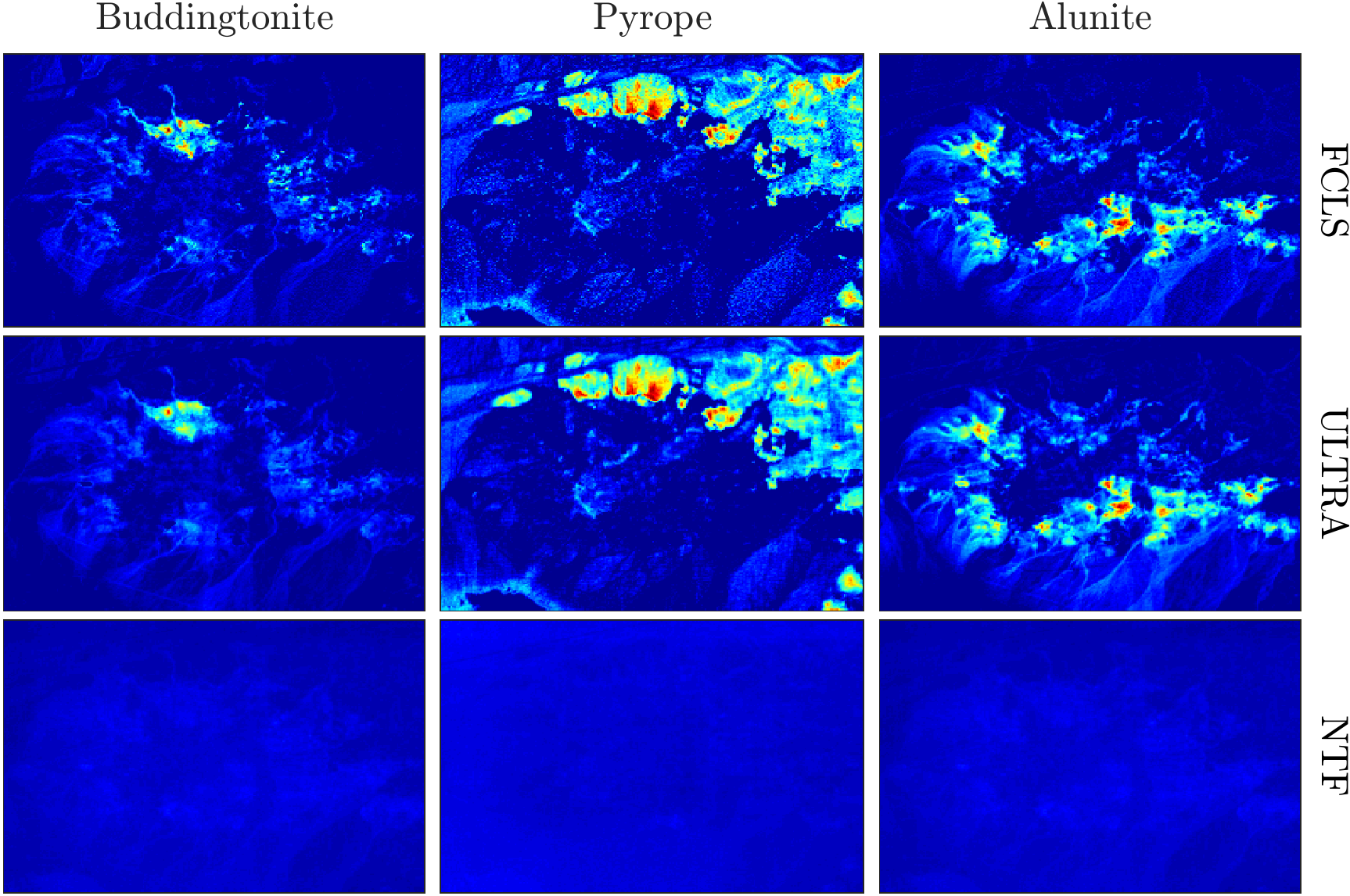}
\caption{Abundance maps of the Cuprite dataset for all tested algotithms where the abundance values are represented by colors ranging from blue ($\alpha_k = 0$) to red ($\alpha_k = 1$).}\label{fig:ab_maps_houston}
\end{figure}

\subsection{Real data}

For simulations with real data we considered the Cuprite Mining Field dataset discussed in~\cite{borsoi2017tech_arxiv}. Figure~\ref{fig:ab_maps_houston} shows the reconstructed abundance maps for all tested methods.
The ULTRA method (middle row) provided accurate abundance estimation that are smoother than the abundances obtained with FCLS. Despite the effort made to tune the parameters, the abundance maps estimated using the NTF method were highly mixed and not coherent with other analyses for this data~\cite{borsoi2017tech_arxiv,Imbiriba2017_bs_tip,Nascimento2005}. In terms of reconstruction error, the FCLS presented the best result (0.0107), which is comparable with the result obtained by the ULTRA method (0.0108), and much smaller than the error obtained by the NTF algorithm (0.0476). When comparing the execution times the FCLS presented the smallest time (1s) followed by the proposed method (45s) and then by the NTF (3459s).



\section{Conclusions}\label{sec:conclusions}

This paper proposed a novel regularization strategy  for linear hyperspectral unmixing. The proposed method imposes an \emph{a priori} low-rank structure to the abundance tensor during the learning process by using a simple regularization that also allows some freedom for the estimated abundances to adequately represent fine scale structures and details.
The proposed strategy is simple and provides accurate results with reasonable increase in the overall problem complexity. In comparison with a recently proposed tensor-based method, the results obtained using the proposed method were significantly better in terms of both accuracy and execution time.

\newpage
\bibliographystyle{IEEEbib}
\bibliography{hyperspectral}

\end{document}